\documentclass{article}
\usepackage[utf8]{inputenc}
\usepackage{graphicx}
\usepackage[margin=1in]{geometry}
\usepackage{amsmath, amssymb}
\usepackage{url}
\usepackage[colorlinks=true,
            linkcolor=blue,
            citecolor=red,
            urlcolor=magenta]{hyperref}

\newcommand{\clip}[3]{\operatorname{Clip}\bigl(#1,#2,#3\bigr)}

\title{Flexible Swarm Learning May Outpace Foundation Models in Essential Tasks}

\author{
    Moein E. Samadi$^{1,2}$, Andreas Schuppert$^{*,1,2}$ \\
    \\
    \small
    $^1$Institute for Computational Biomedicine, RWTH Aachen University, Aachen, Germany. \\
    \small
    $^2$Center for Computational Life Sciences, RWTH Aachen University, Aachen, Germany. \\
    \small
    $^*$\textit{Correspondence:} aschuppert@ukaachen.de
}

\date{}

\begin{document}

\maketitle

\begin{abstract}
Foundation models have rapidly advanced AI, raising the question of whether their decisions will ultimately surpass human strategies in real-world domains. The exponential, and possibly super-exponential, pace of AI development makes such analysis elusive. Nevertheless, many application areas that matter for daily life and society show only modest gains so far; a prominent case is diagnosing and treating dynamically evolving disease in intensive care.

The common challenge is adapting complex systems to dynamic environments. Effective strategies must optimize outcomes in systems composed of strongly interacting functions while avoiding shared side effects; this requires reliable, self-adaptive modeling. These tasks align with building digital twins of highly complex systems whose mechanisms are not fully or quantitatively understood. It is therefore essential to develop methods for self-adapting AI models with minimal data and limited mechanistic knowledge. As this challenge extends beyond medicine, AI should demonstrate clear superiority in these settings before assuming broader decision-making roles.

We identify the curse of dimensionality as a fundamental barrier to efficient self-adaptation and argue that monolithic foundation models face conceptual limits in overcoming it. As an alternative, we propose a decentralized architecture of interacting small agent networks (SANs), inspired by living systems. We focus on agents representing the specialized substructure of the system, where each agent covers only a subset of the full system’s functions. Drawing on mathematical results on the learning behavior of SANs and evidence from existing applications, we argue that swarm-learning in diverse swarms can enable self-adaptive SANs to deliver superior decision-making in dynamic environments compared with monolithic foundation models, though at the cost of reduced reproducibility in detail. 
\end{abstract}

\section{Introduction} 
The debate over the future role of artificial intelligence (AI) is polarized between two extremes: on the one hand, optimistic visions that AI will solve most of humanity’s pressing challenges \cite{kurzweil2005singularity}; on the other, fears that AI may ultimately suppress or even eradicate human society \cite{AI-2027}. In fact, real-world adaptation to dynamic environments, whether opportunities or threats, requires decision-making under uncertainty. 
In this paper, we discuss the capacity of AI to substitute for human decision-making, focusing on strategies of adaptation to dynamic environments. 

We conceptualize humans as open systems composed of mutually interacting entities. Each entity is influenced by external, dynamically varying factors, collectively termed the \textit{environment}, that exhibit non-negligible interactions with the system. For example, in medicine, a patient’s environment includes their broader health status shaped by history, lifestyle, and genetics. 

Moreover, in complex systems, no universal definition of common optimal outcome may exist. The behavior of simple systems, such as prokaryotic cells, can be explained by optimizing the conversion of feedstock into energy. In contrast, mammalian cells integrated within tissues and organs lack such a singular target.

Hence, the goal of the entities within the system is to adapt to external dynamics by pursuing two aims: first, to avoid common negative outcomes, and second, to optimize individual positive outcomes. To achieve these aims, each entity must choose its actions through a three-component decision process:

\begin{enumerate}
    \item[-] \textbf{Sensing} external impacts arising from environmental states on the individual.
    \item[-] \textbf{Prediction} of the outcomes for possible actions within the individual’s portfolio, given its environmental impacts and current system states. This requires a \textit{model} that maps environmental influences and system states to actions and reliably forecasts their consequences. Such predictions must be delivered with acceptable accuracy while consuming minimal resources.
    \item[-] \textbf{Optimization} of action selection through an optimizer that approximates the individual optimum while ensuring avoidance of severe negative outcomes. Short-term gains that almost certainly result in catastrophic long-term consequences are not valid solutions.  
\end{enumerate}

The requirement to consistently avoid severely unfavorable states necessitates sensing systems and models that extend beyond present conditions, anticipating risks of entering harmful states with non-negligible future probability. Thus, the underlying predictive model plays a key role in self-adaptation: it must ensure reliable forecasts under evolving conditions with minimal resource expenditure. 

Such modeling can be approached in two ways. A \textit{foundation model} quantifies the entire space of possible system states, action portfolios, and plausible environmental impacts. Alternatively, a \textit{digital-twin} strategy relies on smaller, incomplete models that represent the local neighborhood of currently active states, combined with continuous adaptation to environmental and system dynamics.  

Recent work in \textit{Nature} has demonstrated that meta-learning across diverse environments can autonomously discover state-of-the-art reinforcement-learning rules (“DiscoRL”), underscoring both the promise of data- and compute-scaled discovery~\cite{oh2025discovering}. This achievement highlights the growing potential of machine-driven discovery but also reinforces the need to explore architectures, such as decentralized swarm learning, that enable scalable self-adaptation by composition rather than monolithic optimization.

In this paper, we first evaluate whether a monolithic, foundation model–based AI could generate models enabling adaption of complex systems to dynamically changing environmental impacts. Foundation models have the indisputable advantage of representing the full body of available knowledge, enabling unprecedented flexibility in mapping diverse external impacts to the complete portfolio of system states and potential actions. They can generate predictions without situation-specific learning, resulting in minimal response times, though at the cost of the enormous investments required for their development. In our discussion, we focus on the conceptual challenges that arise from their universality in the context of dynamic adaptation. 

Second, we discuss a decentralized adaptation strategy inspired by the digital twin concept and motivated by decision-making processes in living systems. Unlike foundation models, this approach relies on small agent networks (SANs) that are continuously updated to reflect specific environmental and system states. We show that SAN architectures combine flexibility with rapid and efficient learning. We argue that networks of small agents representing at least structured knowledge of the underlying functionalities of the system resulting in hybrid (integrated knowledge- and data driven) SAN provide significant advantages compared to unstructured networks of black-box agents. Hence we claim to realize a balanced pay-off between investment of even incomplete knowledge into SANs and gain of performance in adaptation to dynamically varying systems.

Third, we propose a concept of self-adaptive hybrid SANs based on swarm learning. We argue that swarm-learning SANs (SLSANs) in diverse swarms can achieve rapid, efficient, and resilient adaptation to dynamic environments. In an oxygenation case study, we show that SANs restore post phase transition SpO$_2$ forecasting accuracy with approximately 10 minutes of adaptation, whereas a monolithic regressor requires approximately 200 minutes. 

We further propose that SLSANs are conceptually superior to monolithic large-scale models in optimizing adaptation under external stress. Finally, we highlight the critical role of swarm diversity in enhancing resilience and efficiency, while acknowledging the trade-off of reduced reproducibility.

\section{Background and problem setting} 

\subsection{Learning from data with foundation models}

Conceptually, AI-based models are \emph{universal machines}, namely mathematical algorithms that allow:
\begin{itemize}
    \item[-] representing all data of any structure used as inputs for training the machine, by means of any possible function realizable by the algorithm,
    \item[-] discovering and representing all structures within the data that enable mapping a subset of the data to another subset, and
    \item[-] generating new data consistent with the identified structures.
\end{itemize}

The impressive progress of AI in recent years has been driven by a unique combination of technological and scientific innovations in learning concepts, hardware capabilities, and data availability, including:
\begin{itemize}
    \item[-] the availability of GPUs and their large-scale integration into deep neural networks (DNNs),
    \item[-] transformer architectures enabling attention mechanisms beyond purely local features \cite{vaswani2017attention},
    \item[-] efficient backpropagation algorithms for training very large DNNs \cite{rumelhart1986learning},
    \item[-] scaling laws for training large DNNs in language representation, indicating efficiency gains with increasing size \cite{kaplan2020scaling}, and
    \item[-] the availability of very large datasets by automatic searching through the internet.
\end{itemize}

These algorithms enable the representation of extremely large datasets by learning essential and reliable features, thereby compressing data structures beyond lexicographic enumeration as well as interpolating between training samples. Internally, all data structures are represented by vectors in an $n$-dimensional vector space. Crucially, the dimensionality $n$ of data structures in modern AI systems is very high, such that the \emph{curse of dimensionality} poses fundamental challenges:

\begin{enumerate}
    \item[-] \textbf{Exponential data demand:} Learning arbitrary structures in very high dimensions without constraints requires scanning an exhaustive grid, with data demand growing exponentially in $n$. Even with coarse resolution, dimensions on the order of 1000 would require data far exceeding the total content of the internet. The curse of dimensionality can be mitigated only when data exhibit specific structures aligned with specialized algorithms. For example, the surprising success of neural networks in images, text, and time series is mathematically explained by the match between neural architectures and the intrinsic structures of those problems \cite{barron2002universal,mallat2016understanding}. It is evident that the recent progress in language processing and understanding has been primarily driven by the greatly improved alignment between the essential features of language and the capabilities of transformer architectures, compared to unstructured DNNs. In contrast, in several important fields, such as medicine, where AI has not yet fulfilled its promises, the main reason is likely that the relevant features within medical data do not align well with the architectures of currently popular algorithms. This mismatch has led to the surprising observation that some rather traditional algorithms can outperform modern DNN-based approaches \cite{subudhi2021comparing}.

    \item[-] \textbf{Counterintuitive geometry:} In very high dimensions, data distributions tend to concentrate near a surface, leading to counterintuitive phenomena. For instance, in the limit $n \rightarrow \infty $ the distance between any pair of data points tends to become equal. Since inference relies on neighborhoods of the point to be predicted, identifying unique neighborhoods becomes elusive. Moreover, averaging across a dataset often yields results that lie outside the distribution, e.g., a “mean patient” within any (sub-) population may represent no actual patient. Hence, predictive performance in high dimensions depends critically on intrinsic data structures and their compatibility with the algorithm used. 
\end{enumerate}

Importantly, the effects of the curse of dimensionality do not depend on the metrics used to quantify data, nor on preprocessing operations or learning algorithms. These methods merely shift the dimensionality threshold at which reliable learning and prediction fail.

Another fundamental limitation is that, by design, all AI algorithms (as universal machines) cannot \textit{extrapolate}; they can only \textit{interpolate}. Inferring data points outside the set of training data, or extrapolation, is conceptually impossible.  

Whereas in low-dimensional spaces, it is straightforward to determine whether a new data point lies within the interpolation range or whether its prediction requires extrapolation. In very high-dimensional spaces, however, this distinction becomes effectively undecidable. This problem gets even worse if the data structure involves non-continuous data, such as cardinal data structures or even binary data structures \cite{balestriero2021learning}.

One may argue that generative AI appears to extrapolate successfully. However, real-world data are not uniformly distributed; they are clustered, and do not fill the combinatorial space of a hypercube. Hence, in practice, successful applications occur when the point of inference lies within an effectively low-dimensional subspace of the overall combinatorial space, enabling interpolation just in a low-dimensional “local” subspace. For example, generating a video of walking in New York requires interpolation within an effectively low-dimensional subset of all possible videos. Selecting only the available subset of videos and images of New York (or comparable cities) for training restricts the data space to a comparatively low-dimensional subspace, thereby enabling interpolation. In contrast, the notorious hallucinations of generative AI when handling queries for sparsely represented data points may simply result from extrapolation in high-dimensional data spaces without restriction to low-dimensional, densely sampled subspaces.

Another approach to enable extrapolation is the combination of \textit{a priori} knowledge with nearby training features, where the \textit{a priori} knowledge allows the machine to learn structures in the data that can be extended beyond the training domain (zero-shot learning \cite{lampert2013attribute}, ZSL). In ZSL, however, the availability of \textit{a priori} knowledge, or identifiable features within the data, is crucial for successful extrapolation. This \textit{a priori} knowledge can either be provided by humans (in which case the AI can never surpass human expertise, since humans must already know the essential features) or extracted autonomously by the AI from the data itself (where the features are intrinsic to the available training dataset). In the latter case, however, the AI can never be certain that the learned features can be reliably extrapolated beyond the training data, creating a risk of hallucination due to either incorrect feature learning or erroneous assumptions about the generality of the identified features.

One could argue that the emergence of \textit{super-intelligent} systems, AI capable of autonomously improving or re-designing their own algorithms, might be able to overcome this limitation. Essentially, such systems could automatically adapt their algorithms to any features present in the data, thereby achieving an optimal alignment between data structures and algorithmic capabilities. Indeed, the automatic generation of an algorithmic super-structure, functioning as a new form of hyper-data, could represent a breakthrough across many application domains, such as in medicine, where algorithms optimally adapted to specific data structures are not yet known.

However, since the space of possible features within data can be viewed as a combinatorial product space over the data space, the dimensionality of the feature space is dramatically higher than that of the data space itself. In conventional feature learning, the feature space that an AI system can explore is constrained by its intrinsic properties, thereby reducing the effective dimensionality of the search space and enabling established validation strategies to assess trust in the learned features. In the case of a super-intelligent system, however, this constraint would no longer apply, leading to an explosion in the dimensionality of the feature space and raising serious concerns about the feasibility of validation.

Moreover, a super-intelligent system operating solely on the available data would have no means of validating the correctness of the learned features beyond the training dataset. Hence, we do not believe that future super-intelligent systems will be able to overcome the extrapolation problem without the inclusion of \textit{a priori} knowledge, where available.

As a consequence of the general nature of the curse of dimensionality, both the risk of hallucination and the data requirements for training increase exponentially with the dimensionality of the data structures. Given the limited availability of data on the internet, merely adding more of the same data will lead to a saturation phase in which doubling the investment in learning capacity yields only marginal improvements in the overall predictive capability and reliability of AI systems, except in specific applications where the data structures happen to align perfectly with the available algorithms.

\subsection{Foundation models and adaption to dynamic environments}
For quantitative modeling of a rapidly changing system, one solution could be the establishment of a holistic foundation model trained on the complete set of historical and currently available data and knowledge. Under best circumstances, such a system could propose an optimal response to any system's state on short notice, based on all available information. This would significantly outpace human decision-making, which suffers from forgetting and limited learning capacity. Human decisions are naturally based on an incomplete subset of information, often supplemented by irrational intuition and natural stupidity \cite{rich2019lessons}, which can lead to suboptimal or even catastrophic results. 

If the new system state resembles past situations, the model can be based on interpolation from retrospective datasets. Since the likelihood of finding relevant past data increases with the amount of data represented in the decision-making system, a foundation AI model would provide an optimal route to adaptive modeling.  

However, in cases where environmental scenarios the system has to adapt are genuinely new (meaning the combination of environmental settings and system entities differs from all retrospective datasets) decision-making requires extrapolation. This involves making inferences beyond the range of interpolation from historical data.  

For example, in intensive care, a patient’s health status is highly influenced by the functions of multiple, strongly interacting organs that are affected by diverse pathogenic processes, as in Sepsis. Such severe health states are, fortunately, rare over an individual’s lifetime. Consequently, it is unlikely that similar states will recur within the same patient’s history. Because the health state in critical care depends on an extremely high-dimensional set of parameters, good matches with similar patient states, i.e., cases suitable for AI interpolation across the full parameter space, are also rare. The only viable approach is to reduce this full parameter space to a small set of “relevant” parameters that (one hopes) capture the full range of important effects, thereby mitigating the curse of dimensionality. In medical practice, this “breaking of the curse of dimensionality” typically relies on the clinician’s experience. However, persistently high mortality rates and modest prognoses, even among favorable outcomes, indicate a clear need for improvement.

Therefore, directly learning optimal adaptation mechanisms solely from historical data, without disentangling the combinatorics of impact and response, results in sub-optimal outcomes.

\subsection{Knowledge-integrated modeling for self-adaptation}
As discussed above, successful dynamic adaptation of a system to its environment requires models with the capability for limited extrapolation into the neighborhood of the retrospective training data. In particular, the combination of future environmental impacts and system states may not have been observed in the retrospective data, even if both states are at least partially represented individually.

By design, universal machines cannot extrapolate. Therefore, reliable extrapolation must be associated with a loss of universality in the modeling algorithm.

Examples of such restrictions include:
\begin{itemize}
    \item[-] \textbf{Trend estimation beyond the convex hull of the training data:} 
    Trend estimation requires only smoothness of the underlying foundation model. Since smoothness is a generic property in real-world systems, it represents only a minimal restriction of universality in practice. Smoothness-based trend estimation is an established method for extrapolation, particularly when tight monitoring data on system dynamics is available. However, these estimates depend on explicit or implicit estimation of derivatives of the model function, which is an ill-posed problem. As a result, predictions in high dimensions tend to be highly erroneous, restricting the extrapolation range. Importantly, this ill-posedness is algorithm-independent, meaning that trend estimation alone may not be sufficient for self-adaptation strategies in high-dimensional spaces.
    \item[-] \textbf{Integration of explicit \textit{a priori} knowledge into AI systems}, for example through:
    \begin{itemize}
        \item[--] Coupling physics-based equations with AI systems, resulting in the physics-informed neural networks (PINNs) concept \cite{raissi2019physics}.
        \item[--] Reducing dimensionality by model reduction, utilizing intrinsic correlations in the data (e.g., with Autoencoders) or by explicitly integrating prior knowledge into the AI architecture \cite{maier2019learning,samadi2024hybrid,e2025gpt}.
        \item[--] Decomposing the overall system into smaller functional components that interact with each other. Each functionality is represented by a small agent integrated into a network of agents, forming a SAN. The small agent can be implemented either as a black-box model for individual functionalities (each depending only on a small subset of parameters) or as a hybrid model that incorporates structural \textit{a priori} knowledge about the functionality represented. Similar ideas have recently been proposed under the concept of small language models (SLMs) \cite{belcak2025small}. Substituting large, monolithic foundation models with specialized, interconnected functional agents has been demonstrated in the DeepSeek LLM model \cite{dai2024deepseekmoe} and is conceptually related to the “mortal computing” paradigm proposed by G. Hinton \cite{hinton2022forward}.
    \end{itemize}
\end{itemize}
Coupling explicit physics-based equations or integrating physics-based knowledge into PINNs requires human intervention, whereas learning-based model reduction using Autoencoders requires training on the full dataset or explicit prior knowledge. Thus, both approaches appear less suited for naturally occurring rapid self-adaptation in living systems.

The SAN concept, in contrast, requires only a hierarchical decomposition of the overall model into specialized small agents, each representing a real-life functionality, and a network representing the interactions between these agents. The decomposition can be designed according to the available prior knowledge regarding the granularity of functionalities and their interactions. In this approach, knowledge integration is achieved by the decomposition itself, as represented in the interaction network. Since agents are responsible only for specialized tasks rather than full-scale data processing, they can be implemented using small-scale AI systems such as SLMs \cite{belcak2025small}.

Furthermore, because the network topology connects only a limited number of interacting agents, the size of the corresponding adjacency matrix (scaling with $O(n^2_{\text{agent}})$) is orders of magnitude smaller than in full-scale DNNs. Consequently, exchanging and updating the adjacency matrix of the agent-interaction network is far less resource-intensive than updating a full-scale DNN.

A special advantage of the SAN concept, when decomposed according to structural knowledge, derives from prior topological results by Kolmogorov \cite{kolmogorov1961representation} and Arnold on representing high-dimensional functions through functional networks, which later informed the development of Kolmogorov-arnold networks (KANs) \cite{liu2024kan}. Following Kolmogorov and Arnold, Vitushkin demonstrated that interacting networks of small agents operating via smooth functional representations are no longer universal machines \cite{vitushkin1978representation}. The universality of such networks depends on the complexity and smoothness of the small agents, quantified by Vitushkin entropy.

When Vitushkin entropy is low (i.e., when agent complexity is lower than that of the overall model and agent smoothness matches that of the overall model), the SAN ceases to be a universal machine and implicitly encodes structural prior information \cite{samadi2024smooth}. Consequently, SANs can be trained on neighborhoods of low-dimensional subsets of the high-dimensional data space, reducing data requirements for training by orders of magnitude. According to previous work \cite{fiedler2008local}, the dimensionality of the effective data space required for SAN training depends only on the complexity of the small agents. As the granularity level of the decomposition of the overall model can be adapted to the available a priori knowledge, the data demand for training can be controlled by adjusting the level of complexity represented by the small agents in the functional network and their interactions. These results are valid for both discrete and binary data spaces \cite{e2022training}.

However, the trade-off for learning from small datasets and enabling extrapolation is an increase in epistemic uncertainty, particularly concerning the topology of agent interactions and the hybrid model structures within the small agents. In fact, training SANs on noisy data from low-dimensional subsets is an ill-posed problem and requires specialized learning algorithms. Moreover, similar to the extrapolation in PINNs and model-reduction based methods, predictive performance depends on the uncertainty of the interaction topology in the SAN. 

Because of their superior flexibility in functional decomposition (requiring only qualitative prior knowledge) SANs allow optimal design tailored to the specific problem, data availability, extrapolation requirements, and prior knowledge. For these reasons, we propose the SAN concept as the method of choice for realizing self-adaptive systems.

\section{Swarm Learning Strategies for Self-Adaptation}
As discussed above, a strategy for continuous adaptation to a dynamic environment using knowledge-integrated AI technologies requires self-adaptive learning. This involves efficient sensing of external dynamics, identification of epistemic uncertainties in the model, and efficient updating strategies. Because of their extrapolation capabilities and low data demand for training, based only on qualitative \emph{a priori} information, SANs provide a powerful concept for self-adaptive models. As SANs can represent the full information of monolithic foundation models, provided that the data satisfy the constraints derived from the model’s structure \cite{schuppert2011efficient}, they can substitute foundation models within that class of real-life data. Moreover, because of the reduced data requirements for training, detecting emerging changes in the system and adapting to new system states requires less data than in model structures lacking structural \textit{a priori} knowledge. This characteristic provides the potential for rapid self-adaptation to the system’s dynamics.

The price for this advantage, however, is the need to first learn the interaction network (if not known \emph{a priori}) and subsequently re-adapt the network in response to dynamic environmental changes affecting the structure of functional interactions. Identifiability of network structure has been demonstrated for hierarchical decompositions of models into tree structures of functional components \cite{schuppert2011efficient}, but similar results for non-tree structures are lacking. In practice, SAN architectures have been implemented in computational chemical engineering in the context of hybrid models \cite{von2014hybrid}. Here, validation of extrapolation reliability has been performed through model-based design of experiments, using single data samples outside the convex hull of the training set for validation \cite{schuppert2018hybrid}.

Based on these preliminary results, we propose an iterative self-adaptation strategy within swarms of individual SANs. The motivation is twofold: first, the interaction structure is assumed to represent a common framework across agents; second, the individual parameterization of agents depends on heterogeneous optimality targets.  Thus, exchange and mutual learning of the interaction network provide essential information to all individuals, independent of their optimization functions. Because the interaction network is represented by a comparatively small, directed adjacency matrix, it can be exchanged and re-implemented at low resource cost. Because of the fast-learning capabilities of hybrid models, for hybrid-structured agents it is sufficient to exchange the learned optimal hybrid structure instead of the full parameter set of an agent, thereby reducing the required communication resources.

Since targeted design of experiment (as in chemical engineering) is not feasible for self-adaptation in living systems, natural environmental fluctuations (which differently reflected in the parameterizations of SANs within a diverse community) must substitute for designed experiments. For efficient adaptation, it is crucial that epistemic errors caused by mismatches between the interaction structure and external dynamics are recognized as early as possible. Testing the dynamic environment with only a single parametrization is limited to sensing structural aberrations, whereas a swarm of SANs with diverse parametrizations can identify mismatches more effectively. We therefore propose a swarm-learning strategy for adaptation of SAN interaction structures, referred to as SLSAN, which we suggest provides a superior fit to the requirements of self-adaptation in dynamic environments.

The efficiency of SLSAN increases with the number of agents detecting structural mismatches. An effective SLSAN should be based on the following principles:
\begin{itemize}
    \item[-] The swarm consists of diverse agents.
    \item[-] Each individual receives data both from their environment and from other members of the swarm. Swarm members exchange not only information but also decisions derived from it. This information may include detailed process-level data, internal model structure, or aggregated states of each agent. We assume that exchange primarily occurs at the level of model structure and aggregated agent activity.
    \item[-] Each individual processes data to form decisions using a SAN aligned with its optimization target. Agents representing the same functionality operate on the same subset of data but generate outputs depending on their parametrization. These decisions are then aggregated into an overall prediction specific to each SAN.
    \item[-] Individuals exchange both predictions and their fit to reality as sensed individually.
    \item[-] All individuals use the collective information of model fitting across the swarm to perform ensemble-based adaptation of their SAN network structures to systems dynamics.
\end{itemize}

Swarm learning with diverse small agents provides strategic advantages compared to a single monolithic agent:
\begin{itemize}
    \item[-] A single monolithic agent perceives only a single realization of the environmental states and their impact on the system at each point in time. In the case of dynamic behavior of both the environment and the system, such an agent can learn only from the neighborhood of its trajectory. If the learning problem of the overarching interaction structure between the functional modules of the environment is ill-posed with respect to data sampling over a timespan shorter than $T_c$ along the trajectory, the system becomes “blind” after any external change for a duration of $T_c$ following the switch. Since $T_c$ scales exponentially with the complexity of each agent, it can be shortened by reducing agent complexity, though at the cost of greater ambiguity in the estimation of the interaction structure, thereby reducing predictive accuracy. In the case of at least locally ergodic external dynamics, full learning by the agent requires data sampling over the characteristic timespan $T_e$ of the trajectory.

    \item[-] In contrast, a diverse swarm consisting of $m$ agents samples data from multiple starting points in the state space. Consequently, the total learning time $T_e$ can be reduced to $T_e / m$. Moreover, the exchange of information among agents regarding the just the currently learned interaction structure and its associated predictive performance enables the system to perform gradient-based learning within the space of interaction structures between agents. Both effects are expected to significantly reduce the period of “blindness” with respect to external dynamics, thereby enhancing the adaptability of the SLSAN system to dynamic environments. 
\end{itemize}

Overall, we claim that the SLSAN strategy proposed above provides optimal adaptation to systems dynamics. Depending on individual optimization targets, the swarm will exhibit significant diversity, which is essential for efficient adaptation, even if it is suboptimal from the perspective of a common global optimality criterion.

\section{Self-adaptive hierarchical agent networks enable robust oxygenation forecasting}
As an example for the SLSAN strategy described above, we introduce a \emph{Self-Adaptive Hierarchical Agent Network} (SAHA-Net) for one-step-ahead forecasting of arterial oxygen saturation (SpO\textsubscript{2}) across regime shifts in pulmonary physiology in intensive care. Our example is motivated by the needs of intensive care which are paradigmatic for a wide area in medical application areas. SAHA-Net is evaluated using a digital twin of gas exchange (Appendix~\ref{app:A}) and compared against a monolithic gradient-boosting regressor (Appendix~\ref{app:B}). 

The digital twin generates synthetic data using classic gas exchange relationships (alveolar ventilation, alveolar gas equation, shunt and alveolar–arterial (A--a) gradient) and an observation process that mimics sensor inertia and noise. At the midpoint $t^\star{=}360$, the physiology undergoes a “phase transition” to an acute respiratory distress syndrome (ARDS)-like state (Appendix~\ref{app:A}). 
The forecasting target is the observed oxygen saturation, denoted by 
$\text{SpO}_2^{\mathrm{obs}}$. 
The input vector at time $t$, $u_t \in \mathbb{R}^d$, consists of ventilator 
settings, posture, compliance, and derived physiological features 
(Eq.~\eqref{eq:inputs}).

\subsection{Architecture of SAHA-Net}
SAHA-Net embodies our SLSAN hypothesis: by partitioning features across small, specialized agents in a knowledge-driven manner and coordinating them through supervised communication, the architecture supports flexible reconfiguration of agent structure. This reduces the effective parameter complexity and prepares the system for adaptive responses to environment-induced changes via a swarm learning strategy. 

Concretely, in the first level of SAHA-Net hierarchy designed for oxygenation forecasting, three agents focus on physiologically distinct mechanisms, namely ventilation/dead space ($\mathcal{A}_V$), A--a gradient ($\mathcal{A}_G$), and shunt ($\mathcal{A}_S$), while a second-level supervisor $\mathcal{S}$ manages their communication and fuses their forecasts towards the model's final outcome (Appendix~\ref{app:C}). 

At each minute $t$, agents observe a masked subset 
$ M_i\in\{0,1\}^{d}$ of the input vector, where $i\in\{V,G,S\}$, and generate private one-step forecasts from their masked features (Eq.~\eqref{eq:firstpass}). Agents then share their private forecasts over a directed link matrix, $C\in\{0,1\}^{3\times 3}$ (where $C_{ii}=0$ and $C_{j\to i}=1$ permits sending message from $\mathcal{A}_j$ to $\mathcal{A}_i$), controlled by the supervisor (Eq.~\eqref{eq:neighbor-mean}). Each agent updates its belief by convexly blending its private forecast with the neighbor aggregate (Eq.~\eqref{eq:convex-update}). Next, the supervisor receives the three updated agent forecasts as a feature vector and produces the final prediction (Eq.~\eqref{eq:superfusion}). 
This architecture promotes modular specialization and low dimensional communication among agents, features we hypothesize may help overcome the limitations of foundation models in dynamic environments.

\subsection{Self-adaptation through swarm learning}
Before the phase transition time $t^\star$, we use prior physiological knowledge to initialize $\{M_i^{\text{prior}}, C^{\text{prior}}\}$ without hard-wiring them. 
After the phase transition time $t^\star$, the network not only updates the weights of the regressors associated to the agents and the supervisor, but also adapts its \emph{structure}. We define the structural degrees of freedom as
\[
\theta=\big(M_V,M_G,M_S,\,C\big),\quad M_i\in\{0,1\}^{d},\; C\in\{0,1\}^{3\times 3},\; C_{ii}=0,
\]
and optimize $\theta$ on a short adaptation window $T_{\mathrm{adapt}}=[t^\star, t_{\text{train/test split}}]$ using particle swarm optimization (PSO) restricted to these discrete variables (Appendix~\ref{app:C}).

\begin{figure}[h]
    \centering
    \includegraphics[scale=0.064]{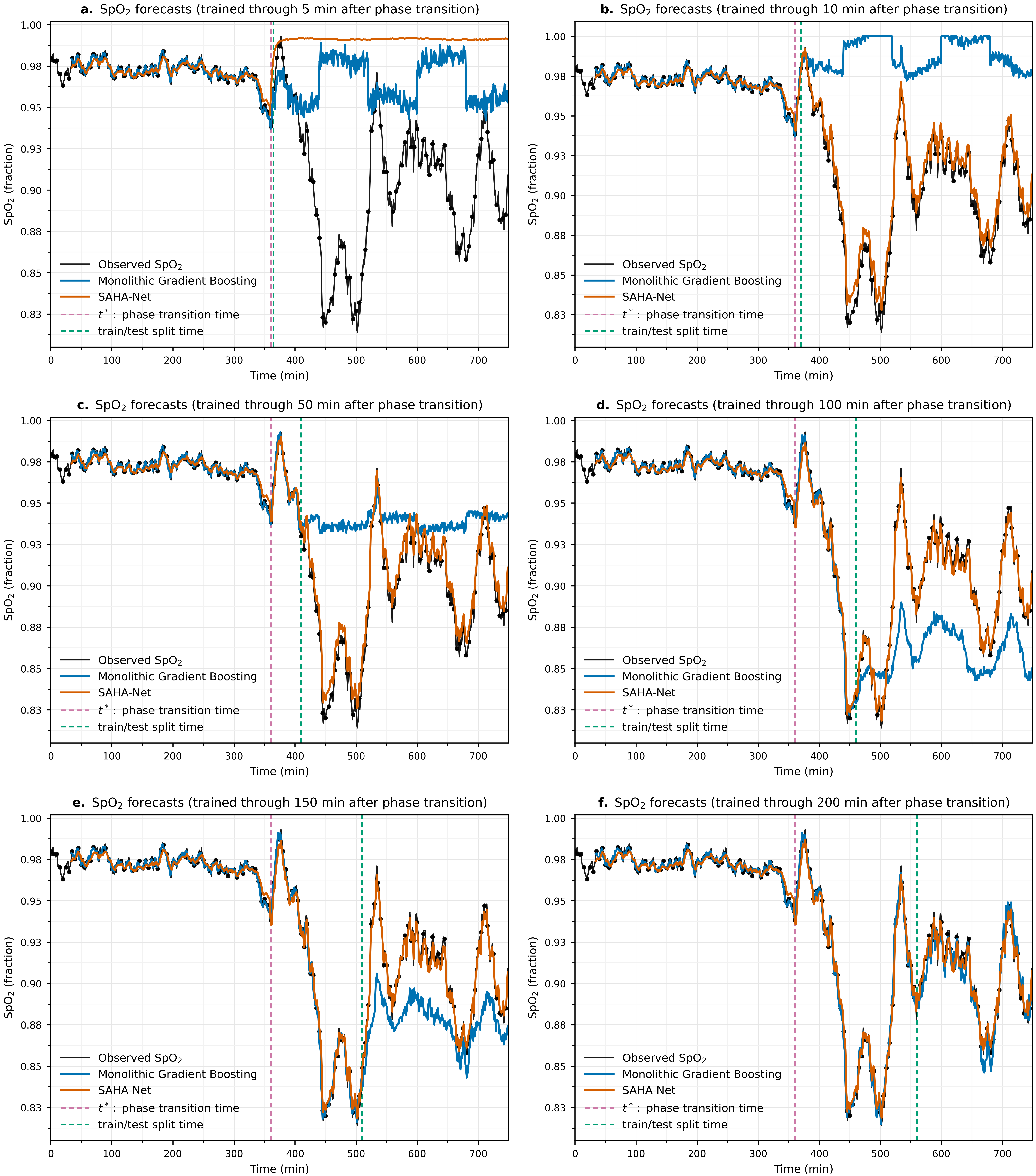}
    \caption{One-step-ahead \(\text{SpO}_2\) forecasts across a phase transition. Each panel uses a different adaptation horizon after the transition 
    (\(T_{\mathrm{adapt}}\in\{5,10,50,100,150,200\}\) minutes).
    SAHA-Net rapidly aligns with post-transition oxygenation dynamics, while the monolithic model adapts more slowly, requiring longer horizons.}
    \label{fig:forecast_results}
\end{figure}

Figure~\ref{fig:forecast_results} compares the monolithic gradient-boosting regressor and SAHA-Net for one-step-ahead \(\text{SpO}_2\) forecasting across the induced phase transition at \(t^\star{=}360\), with panels corresponding to post-transition adaptation windows \(T_{\mathrm{adapt}}\in\{5,10,50,100,150,200\}\) minutes.

Immediately after the transition, with only 5 minutes of post-transition data, both models degrade and fail to track the new dynamics. As the adaptation window increases to 10 minutes, SAHA-Net adapts its structure and rapidly recovers, whereas the monolithic regressor remains biased toward the pre-transition regime, producing over-smoothed trajectories through the first 50 minutes. By 100--150 minutes of adaptation, the monolithic model still underreacts and lags. When \(T_{\mathrm{adapt}}=200\) minutes, the monolithic model also closely follows the amplitude and timing of the post-transition dynamics, reproducing both sharp drops and partial recoveries observed in \(\text{SpO}_2^{\mathrm{obs}}\).

As shown in Figure~\ref{fig:forecast_results}, SAHA-Net rapidly adapts to post-transition oxygenation dynamics, whereas the monolithic gradient-boosting regressor adapts more slowly and requires longer adaptation windows \(T_{\mathrm{adapt}}\). This supports the hypothesis that structural self-adaptation via agent masks and inter-agent communication reduces the amount of post-transition data required to recover forecasting accuracy under environment-induced changes in system dynamics, here mimicked by an ARDS-like phase transition in oxygenation.

\section{Conclusion}

In our paper, we discussed the drawbacks of monolithic foundation models arising from the curse of dimensionality, a challenge inherent to all high-dimensional systems. We argued that the curse of dimensionality poses unresolved challenges for monolithic foundation models in tasks that require decision-making in response to the behavior of complex systems under dynamic environments. Due to the inherent properties of high-dimensional systems and environments, there is evidence that large foundation models are conceptually not well-suited to solving such tasks.

As an alternative, we propose the concept of swarm learning through networks of small agents. Conceptually, such networks exhibit mathematical features that are complementary to those of monolithic foundation models and enable rapid and efficient adaptation to external dynamics. Moreover, they allow systematic and unrestricted integration of \emph{a priori} knowledge at all levels, thereby improving the efficiency of learning. As a result, small agent networks can be flexibly adapted to the availability of \emph{a priori} knowledge, data, and performance requirements.

As an alternative, we propose the concept of swarm learning through networks of small hybrid agents. Conceptually, such networks exhibit mathematical features that are complementary to those of monolithic foundation models and enable rapid and efficient adaptation to external dynamics. Moreover, they allow systematic and unrestricted integration of \textit{a priori} knowledge at all, even just qualitative levels, thereby improving the efficiency of learning. As a result, small agent networks can be flexibly adapted to the availability of \textit{a priori} knowledge, data, and performance requirements.

\clearpage



\section*{Appendices}
\appendix
\section{Digital Twin of Oxygenation Dynamics: Synthetic Physiology and Observation} 
\label{app:A} 
\setcounter{equation}{0}  
\renewcommand{\theequation}{A.\arabic{equation}} 
We constructed a digital twin of pulmonary gas exchange, evolving over discrete minutes 
$t=1, \dots ,N$ with $N= 720$. At the halfway point $t^\star= 360$, a regime change (“phase transition”) in lung physiology is triggered. 

At each time $t$ the recorded parameters are inspired oxygen fraction (FiO$_2$), 
positive end-expiratory pressure (PEEP), tidal volume (VT, mL), respiratory rate 
(RR, breaths/min), a proning indicator, and lung compliance 
(CL, mL/cmH$_2$O). Constants are fixed as
\begin{align*}
P_b = 760 \; \text{(mmHg)},\quad P_{H_2O}= 47 \; \text{(mmHg)},\quad
R=0.8,\quad V_{\text{CO}_2}= 200 \; \text{(mL/min)},\quad 
\text{SvO}_2=0.70,
\end{align*}
together with a Hill dissociation curve defined by $P50= 26.6 \; \text{(mmHg)}$  and Hill coefficient $n=2.7$, predicted body weight $\text{PBW}= 70 \; \text{(Kg)}$, and anatomic  dead space $\text{VD}_{\text{anat}}=2.2\cdot\text{PBW}$ (mL).

\paragraph{Regime schedule.}  
A logistic transition variable increases sigmoidally after $t^\star$:  
\begin{align}
    g(t)=\frac{1}{1+\exp\!\bigl(-\tfrac{t-t^\star}{\tau_g}\bigr)},  
    \quad \tau_g\in[12,20]\text{ minutes (case-specific).} 
\end{align}  
This regime modulates several parameters: compliance falls, alveolar dead space and shunt
rise, and the A–a gradient worsens. PEEP and proning try to mitigate these effects.

\paragraph{Alveolar ventilation and PaCO$_2$.}
Auto-PEEP fraction $\phi_{\text{auto}}$ increases with tachypnea and low compliance:  
\begin{align}  
\phi_{\text{auto}}(t)=\clip{k_{RR}\bigl(RR(t)-14\bigr) + k_{CL}\bigl(35-CL(t)\bigr)}{0}{0.08},  
\label{A2}
\end{align}  
with typical $k_{RR}=0.002$, $k_{CL}=0.004$ pre-transition, and the clipping operator $\clip{x}{a}{b}$ restricting a value $x$ to the interval $[a,b]$ 
$$
\clip{x}{a}{b} = \min(\max(x,a),b).
$$
Then, the effective and alveolar tidal volumes are:  
\begin{align}  
\text{VT}_{\text{eff}}(t)=\text{VT}(t)\bigl(1-\phi_{\text{auto}}(t)\bigr),\quad  
\text{VT}_{\text{alv}}(t)=\max\bigl(\text{VT}_{\text{eff}}(t)-\text{VD}_{\text{anat}}-f_{DS}(t)\,\text{VT}_{\text{eff}}(t),\,5\bigr), 
\label{A3}
\end{align}  
where alveolar dead-space fraction depends on regime $g$ and PEEP  
\begin{align}  
f_{DS}(t)=\clip{0.10\bigl(1-g(t)\bigr)+0.25\,g(t)-0.005\,g(t)\,\bigl(\text{PEEP}(t)-5\bigr)}{0.05}{0.35}.  
\label{A4}
\end{align}  
Alveolar ventilation (L/min) and arterial CO$_2$ follow:
\begin{align}
V_A(t)=\frac{\text{VT}_{\text{alv}}(t)\cdot RR(t)}{1000},\qquad
\text{PaCO}_2(t)=\clip{\frac{863\,V_{\text{CO}_2}}{\max(V_A(t),\,0.5)}}{25}{80}.
\label{A5}
\end{align}

\paragraph{Alveolar O$_2$ and A--a gradient.} 
Alveolar oxygen tension derives from the alveolar gas equation:
\begin{align}
P_{A}O_2(t)=\text{FiO}_2(t)\bigl(P_b - P_{H_2O}\bigr)-\frac{\text{PaCO}_2(t)}{R}.
\end{align}
A--a gradient increases with regime and is mitigated by PEEP and proning:  
\begin{align}  
A\!-\!a(t)=\clip{\underbrace{10\bigl(1-g(t)\bigr)+45\,g(t)}_{\text{regime baseline}}  
-2\,g(t)\bigl(\text{PEEP}(t)-5\bigr)-4\,g(t)\cdot\text{Prone}(t)+\varepsilon_{A\!-\!a}(t)}{5}{80}.  
\end{align}  
Capillary oxygen partial pressure is then:
\begin{align}
P_aO_2^{\text{cap}}(t)=\clip{P_{A}O_2(t)-A\!-\!a(t)}{30}{600}.
\end{align}

\paragraph{Shunt and arterial saturation.}  
The shunt fraction rises with regime but is mitigated by PEEP and proning:
\begin{align}
\text{shunt}(t)=\clip{  
\underbrace{0.05\bigl(1-g(t)\bigr)+0.32\,g(t)}_{\text{regime baseline}}  
-0.015\,g(t)\bigl(\text{PEEP}(t)-5\bigr)-0.06\,g(t)\,\text{Prone}(t)+\varepsilon_s(t)}{0.02}{0.45}.  
\end{align}  
Capillary oxygen saturation is given by the Hill curve:
\begin{align}
S_{a}O_2^{\text{cap}}(t)=\frac{P_aO_2^{\text{cap}}(t)^{n}}{P_aO_2^{\text{cap}}(t)^{n}+P50^{n}}.
\end{align}
Finally, the true arterial saturation combines capillary and venous blood according 
to the shunt fraction:
\begin{align}
\boxed{\,S_{a}O_2(t)=\clip{(1-\text{shunt}(t))\,S_{a}O_2^{\text{cap}}(t)+\text{shunt}(t)\cdot\text{SvO}_2}{0.5}{1.0}
=\text{SpO}_{2,\text{true}}(t).\;}
\end{align}

\paragraph{Observation model.} 
Pulse oximetry does not measure $S_{a}O_2(t)$ instantaneously. Instead, the  measured saturation $\text{SpO}_2^{\text{obs}}(t)$ is modeled as a  low-pass filtered and noisy version of $\text{SpO}_{2,\text{true}}(t)$.

\textit{Inertial dynamics.} We introduce an exponentially weighted moving average with time constant 
$\tau_{\text{sens}}=10$ seconds. In discrete time this is
\begin{align}
\tilde{S}(t)=\alpha\,\tilde{S}(t-1)+(1-\alpha)\,S_{a}O_2(t),
\qquad
\alpha=\exp\!\left(-\tfrac{\Delta t}{\tau_{\text{sens}}}\right),
\end{align}
with $\Delta t= 60 (s)$ per time step. This smooths sudden changes 
in $S_{a}O_2(t)$.

\textit{Measurement noise.} Additive Gaussian noise $\varepsilon(t)\sim\mathcal{N}(0,\sigma^2)$ with 
$\sigma=0.01$ models sensor variability and motion artefacts. 
The final observed saturation is then
\begin{align}
\boxed{\;\text{SpO}_2^{\text{obs}}(t)=
\clip{\tilde{S}(t)+\varepsilon(t)}{0.5}{1.0}.\;}
\end{align}

\paragraph{Phase transition (onset of ARDS-like physiology).}  
The transition is realized as a coordinated change in hidden and semi-hidden physiology driven by $g(t)$:  
\begin{itemize}  
    \item[-] Compliance CL$(t)$ decreases with $g(t)$ (stiffer lungs).  
    \item[-] Alveolar dead space fraction $f_{DS}(t)$ increases with $g(t)$, mildly improved by higher PEEP.  
    \item[-] A--a gradient increases with $g(t)$ (worsened V/Q), mitigated by PEEP and proning.  
    \item[-] Shunt increases with $g(t)$ (recruitment failure), mitigated by PEEP and proning.  
    \item[-] Auto-PEEP fraction grows due to tachypnea and low compliance, reducing $V_A$ and raising PaCO$_2$.  
\end{itemize}

\section{Monolithic Gradient Boosting Regressor Model} 
\label{app:B}     
\setcounter{equation}{0}  
\renewcommand{\theequation}{B.\arabic{equation}} 
The monolithic gradient boosting regressor model is trained to predict the next-step oxygen saturation $\hat y_{t+1\mid t}$ from a feature vector $\mathbf{x}_t$. 
The predictor is a gradient-boosted decision tree regressor:
\[
  \hat{y}_{t+1\mid t}^{\text{mono}} = f^{\text{mono}}(\mathbf{x}_t),
\]
where $f^{\text{mono}}$ is trained using the feature matrix $X_t$ and targets $y_t$. 

\paragraph{Feature set $X_t$.}  
The feature matrix for each time $X_t$ consists of 13 features (see Appendix \ref{app:A} for more details):
\begin{itemize}  
\item[-] \textbf{Six raw input features:} FiO$_2$, PEEP, VT (mL), RR, Prone, and CL (mL/cmH$_2$O).  
\item[-] \textbf{Three derived physiology features:} $\text{PaCO}_2(t)$ and $\text{V}_A(t)$ computed from the digital twin formulas using observed inputs (see Eq.s \ref{A2}--\ref{A5}), and $\text{P}_A\text{O}_2(t)$:
\begin{align}
  \text{P}_A\text{O}_2(t) &= \mathrm{FiO_2}_{,t}\cdot(P_b-P_{\mathrm{H_2O}}) - \frac{\mathrm{PaCO_2}_{,t}}{R}.
\end{align}

\item[-] \textbf{Four input delta features:} $\Delta\text{FiO}_2$, $\Delta\text{PEEP}$, $\Delta\text{RR}$, $\Delta\text{Prone}$.  \\
One-step deltas are
\begin{align*}
  & \Delta \mathrm{FiO_2}_{,t} = \mathrm{FiO_2}_{,t} - \mathrm{FiO_2}_{, t-1},\\
  & \Delta \mathrm{PEEP}_t = \mathrm{PEEP}_t - \mathrm{PEEP}_{t-1}\\
  & \Delta \mathrm{RR}_t = \mathrm{RR}_t - \mathrm{RR}_{t-1}\\
  & \Delta \mathrm{Prone}_t = \mathrm{Prone}_t - \mathrm{Prone}_{t-1}.
\end{align*}
\end{itemize}  

\paragraph{Training and early stopping.}  
The model is trained on pre-transition regions, split into a training prefix and a pre-validation tail. 
Hyperparameter tuning is performed via a grid search with early sopping of 10 rounds over learning rate, maximum tree depth, minimum samples per leaf, and subsampling fraction.

For each configuration, staged prediction on the pre-validation tail determines the optimal number of trees by minimizing the one-step-ahead mean squared error (MSE). 
The final model is then refit on the full pre-training window using the selected number of trees. 
The chosen hyperparameters are:
\[
  \text{\#n\_estimators}=1200,\quad 
  \text{learning\_rate}=0.01,\quad 
  \text{max\_depth}=3,\quad 
  \text{subsample}=0.7,\quad 
  \text{min\_samples\_leaf}=8.
\]

\section{Self-Adaptive Hierarchical Agent Network (SAHA-Net)} 
\label{app:C}                            
\setcounter{equation}{0}  
\renewcommand{\theequation}{C.\arabic{equation}} 
We present a hierarchical multi-agent model for one-step-ahead SpO\textsubscript{2} forecasting. 
The model observes SpO\textsubscript{2} at discrete times $t=1, \dots, N$ and seeks the
one-step-ahead forecast $\hat y_{t+1\mid t}$ from information up to time $t$.

\paragraph{Input vector.}  
Each time step exposes the following input vector:
\begin{equation}
u_t=\big\{
\mathrm{FiO}_2, \mathrm{PEEP}, \mathrm{RR}, \mathrm{VT}, \mathrm{CL},\ \mathrm{Prone},
\mathrm{PaCO}_2, \mathrm{P_AO}_2, \mathrm{V_A},\
\Delta \mathrm{FiO}_2,\ \Delta \mathrm{PEEP}, \Delta \mathrm{RR}, \Delta \mathrm{Prone}
\big\}_t .\label{eq:inputs}
\end{equation}

\paragraph{Model hierarchy.} 
The model is organized as a two-level hierarchy: in the first level, three physiological agents $\mathcal{A}_V$ (ventilation/dead space), $\mathcal{A}_G$ (A–a gradient), and $\mathcal{A}_S$ (shunt) receive inputs, and in the second level, a single supervisor agent $\mathcal{S}$ manages their connections and inputs to produce the final prediction.

At the first level of hierarchy, each of the three agents is a gradient boosting regressor that uses masked $u_t$ to forecast one-step-ahead SpO\textsubscript{2}:

\begin{itemize}
    \item[-] \textbf{$\mathcal{A}_V$ (ventilation/dead space agent):} 
    this agent tends to favor $\mathrm{PEEP}, \mathrm{RR}, \mathrm{VT}, \mathrm{CL}$, and their changes, plus $\mathrm{V_A}$.
    \item[-] \textbf{$\mathcal{A}_G$ (A–a gradient agent):} 
    this agent tends to favor $\mathrm{FiO}_2, \mathrm{PEEP}, \mathrm{Prone}, \mathrm{P_AO}_2$, and $\Delta$ terms, plus $\mathrm{PaCO}_2$ and $\mathrm{P_AO}_2$.
    \item[-] \textbf{$\mathcal{A}_S$ (shunt agent):} 
    like the gradient agent, this agent tends to favor $\mathrm{FiO}_2, \mathrm{PEEP}, \mathrm{Prone}, \mathrm{P_AO}_2$, and $\Delta$ terms, plus $\mathrm{PaCO}_2$ and $\mathrm{P_AO}_2$.
\end{itemize}
These “tendencies” are not hard-coded. Instead, a flexible mask dynamically adapts them, activating or deactivating specific input$\rightarrow$agent links. 

The agents then share their forecasts with one another. A second-level supervisor controls this communication: 
\begin{itemize}
    \item[-] \textbf{$\mathcal{S}$ (supervisor agent):} 
    the supervisor determines which inputs each agent receives, how they communicate, and how their outputs are fused towards the final prediction.
\end{itemize}

\paragraph{Structuring hierarchical multi-agent networks.}
Let $d$ denote the dimension of $u_t$ in \eqref{eq:inputs}. For each agent $\{ \mathcal{A}_i \; | \;  i\in\{V,G,S\} \}$,
let $M_i\in\{0,1\}^{d}$ be a binary mask that selects which coordinates of $u_t$
are visible to that agent, and let $z^{(i)}_t := M_i \odot u_t$ denote the masked
input (Hadamard product). Each agent produces a private one–step–ahead forecast
from only its masked inputs:
\begin{equation}
\hat y^{(i,1)}_{t+1\mid t} \;=\; f^{\mathcal{A}_i}\!\big(z^{(i)}_t;\,\beta_i\big),
\qquad i\in\{V,G,S\}, \label{eq:firstpass}
\end{equation}
where $f_i(\cdot;\beta_i)$ is a gradient boosting regressor whose parameters $\beta_i$ are fit on the current training/validation window.

Furthermore, agents exchange their private predictions. The supervisor maintains
a binary link matrix $C\in\{0,1\}^{3\times 3}$ with $C_{ii}=0$; $C_{j\to i}=1$
permits sending $\hat y^{(j,1)}_{t+1\mid t}$ from $j$ to $i$. For receiver $i$,
define the neighbor set $\mathcal{N}_i:=\{j\neq i:\ C_{j\to i}=1\}$. The neighbor
aggregate is the uniform mean of permitted senders’ private predictions:
\begin{equation}
\bar y^{(-i)}_{t+1\mid t} \;=\;
\begin{cases}
\displaystyle \frac{1}{|\mathcal{N}_i|}\sum_{j\in\mathcal{N}_i}\hat y^{(j,1)}_{t+1\mid t}, & |\mathcal{N}_i|>0,\\[8pt]
\hat y^{(i,1)}_{t+1\mid t}, & \text{otherwise.}
\end{cases}
\label{eq:neighbor-mean}
\end{equation}

Therefore, the variable set 
\[
\theta \;=\; \big(M_V,M_G,M_S,\ C\big),\quad
M_i\in\{0,1\}^{d},\ \ C\in\{0,1\}^{3\times 3},\ C_{ii}=0.
\] 
defines the structure of the hierarchical multi-agent network. This parameter set $\theta$ is then optimized via a swarm learning strategy.

\paragraph{Updated agent forecasts and supervisor fusion.}
Agent $i$ updates its forecast by convexly
combining its own private prediction with the neighbor aggregate:
\begin{equation}
\hat y^{(i,2)}_{t+1\mid t} \;=\;
(1-\lambda_i)\,\hat y^{(i,1)}_{t+1\mid t}
\;+\; \lambda_i\, \bar y^{(-i)}_{t+1\mid t},
\qquad \lambda_i\in[0,1),
\label{eq:convex-update}
\end{equation}
where $\lambda_i$ is estimated on the current training/adaptation window by
least squares with projection onto $[0,1)$. Note that $\lambda$ is not allowed to become 1. This constraint prevents the agent from effectively self-terminating in an update.

The supervisor constructs a fusion feature vector from the three updated agent
predictions and feeds it to a gradient boosting regressor:
\begin{equation}
x^{\mathrm{sup}}_t :=
\begin{bmatrix}
\hat y^{(V,2)}_{t+1\mid t} \\
\hat y^{(G,2)}_{t+1\mid t} \\
\hat y^{(S,2)}_{t+1\mid t}
\end{bmatrix},
\qquad
\hat y_{t+1\mid t} \;=\; f^{\mathcal{S}}\!\left(x^{\mathrm{sup}}_t;\,\psi\right), \label{eq:superfusion}
\end{equation}
where $\psi$ are fixed hyperparameters trained on the current window.

\paragraph{Self-adaptive model updates via swarm learning.} 
Let $t^*$ denote the onset of the phase transition, or the switch time. For the pre-switch period ($t < t^*$), we employ the prior knowledge-based structure $\{M_i^{\mathrm{prior}}, C^{\mathrm{prior}}\}$, which captures the main feature tendencies of the agents with minimal links between them, and fit all continuous parameters on the pre-switch window. For the post-switch period ($t \geq t^*$), we define a short adaptation window between the switch time and the train/test split point, $T_{\mathrm{adapt}} = [t^*, t_{\mathrm{train/test \; split}}]$, during which we adapt the hierarchical structure of the multi-agent system (i.e., the inputs each agent receives and the allowed links) while simultaneously refitting the agent regressors $f^{\mathcal{A}_i}$, the convex weights $\lambda_i$, and the supervisor $f^{\mathcal{S}}$.

We then restrict particle swarm optimization to structural variables only:
\[
\theta \;=\; \big(M_V,M_G,M_S,\ C\big),\qquad
M_i\in\{0,1\}^{d},\ \ C\in\{0,1\}^{3\times 3},\ C_{ii}=0.
\]
For any candidate $\theta$, we:
\begin{enumerate}
  \item[-] fit $f_i(\cdot;\beta_i)$ and compute private forecasts \eqref{eq:firstpass};
  \item[-] form neighbor aggregates \eqref{eq:neighbor-mean} and update agents via
        \eqref{eq:convex-update} (fit $\lambda_i\in[0,1)$ by least squares);
  \item[-] train the supervisor $f^{\mathcal{S}}$ to produce \eqref{eq:superfusion}.
\end{enumerate}
The fitness on $T_{\mathrm{adapt}}$ is
\begin{equation}
J(\theta) \;=\;
\frac{1}{|T_{\mathrm{adapt}}|}\sum_{t\in\mathcal{T}_{\mathrm{adapt}}}
\big(y_{t+1}-\hat y_{t+1\mid t}\big)^2
\;+\;\lambda_M \sum_{i\in\{V,G,S\}} \|M_i\|_0
\;+\;\lambda_C \,\|C\|_0,
\label{eq:fitness}
\end{equation}
i.e., mean squared error plus small penalties on the number of selected inputs
per agent and the number of active links. We run a modest swarm (30
particles for 30 iterations); particle positions are thresholded to
$\{0,1\}$ at evaluation time, continuous parameters are refit, and $J(\theta)$
is evaluated with best updates.

\end{document}